\documentclass[runningheads]{llncs}
\usepackage{graphicx}

\usepackage{tikz}
\usepackage{comment}
\usepackage{amsmath,amssymb} 

\usepackage{multirow}
\usepackage{xspace}

\usepackage[width=122mm,left=12mm,paperwidth=146mm,height=193mm,top=12mm,paperheight=217mm]{geometry}

\newcommand{\etal}{\textit{et al}.}
\newcommand{\ie}{\textit{i}.\textit{e}.\@\xspace}
\newcommand{\eg}{\textit{e}.\textit{g}.\@\xspace}

\begin{document}
\pagestyle{headings}
\mainmatter

\title{CrowdMLP: Weakly-Supervised Crowd Counting via Multi-Granularity MLP} 

\author{Mingjie Wang, Jun Zhou, Hao Cai, Minglun Gong}
\institute{University of Guelph, ON, Canada\\
	Memorial University of Newfoundland, NL, Canada\\
	Dalian Maritime University, Dalian, China}

\maketitle

\begin{abstract}
Existing state-of-the-art crowd counting algorithms rely excessively on location-level annotations, which are burdensome to acquire.  When only count-level (weak) supervisory signals are available, it is arduous and error-prone to regress total counts due to the lack of explicit spatial constraints.  To address this issue, a novel and efficient counter (referred to as CrowdMLP) is presented, which probes into modelling global dependencies of embeddings and regressing total counts by devising a multi-granularity MLP regressor.  In specific, a locally-focused pre-trained frontend is cascaded to extract crude feature maps with intrinsic spatial cues, which prevent the model from collapsing into trivial outcomes. The crude embeddings, along with raw crowd scenes, are tokenized at different granularity levels.  The multi-granularity MLP then proceeds to mix tokens at the dimensions of cardinality, channel, and spatial for mining global information.  An effective proxy task, namely Split-Counting, is also proposed to evade the barrier of limited samples and the shortage of spatial hints in a self-supervised manner.  Extensive experiments demonstrate that CrowdMLP significantly outperforms existing weakly-supervised counting algorithms and performs on par with state-of-the-art location-level supervised approaches.

\end{abstract}

\section{Introduction}

Object counting strives to estimate the total number of objects dispersing across still images~\cite{song2021rethinking} or dynamic video sequences~\cite{ma2021spatiotemporal}. It has increasingly drawn attention from computer vision community, thanks to its wide spread societal applications, e.g., social distance monitoring~\cite{punn2020monitoring}, traffic surveillance~\cite{Zhang_2017_ICCV}, counting in agriculture~\cite{lu2017tasselnet} and metropolis crowd management~\cite{ma2021towards}.

\begin{figure}[t]
	\begin{center}
		\includegraphics[width=0.6\linewidth]{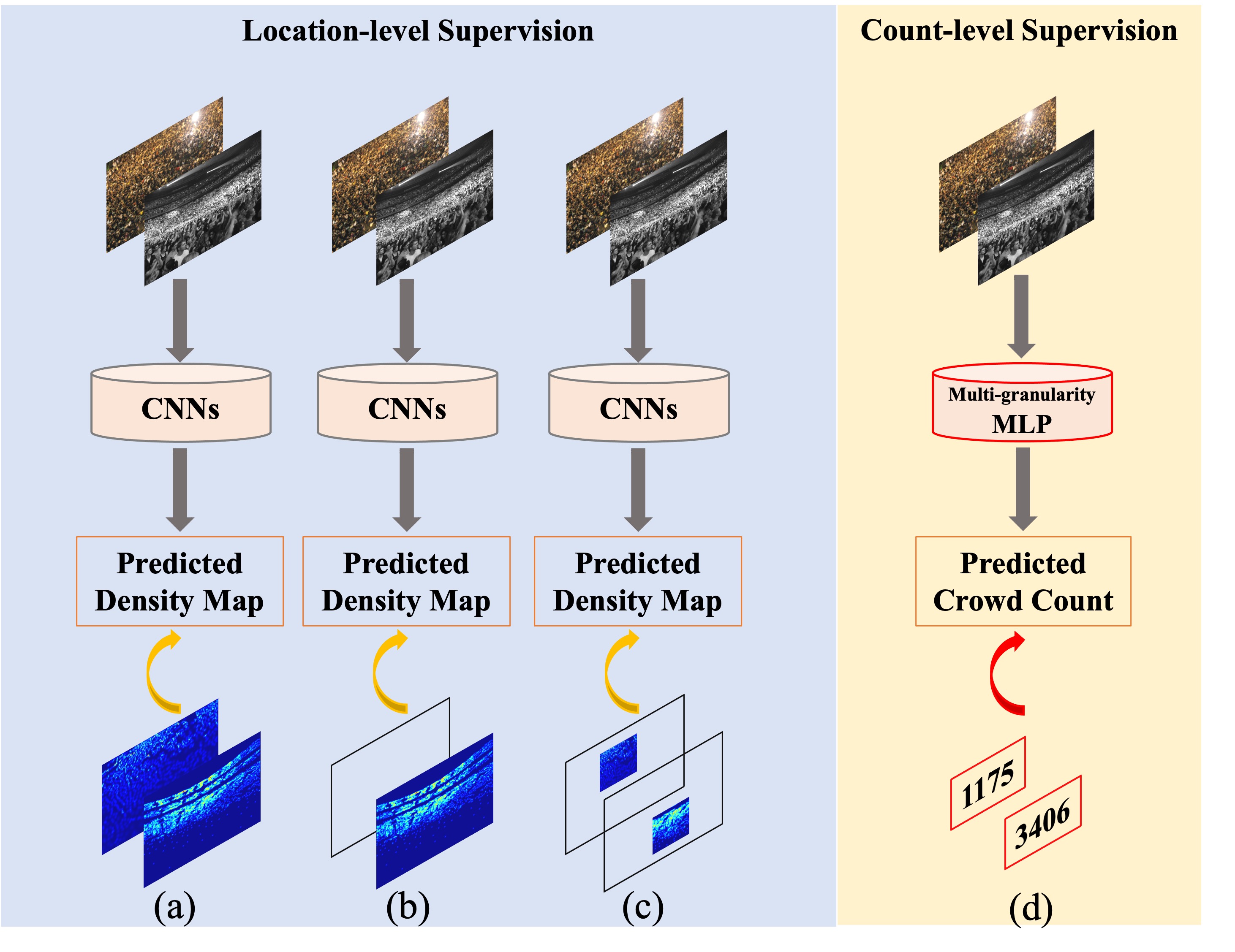}
	\end{center}
	\vspace{-10pt}
	\caption{The categories of crowd counter under distinct supervisory signals. (a) Learn from an entire set of location-level density maps. (b) Learn from density maps of partially selected images. (c) Learn from density maps of partial regions in crowd scenes. (d) Our approach learns directly from less time-consuming  count-level (weak) annotations.}
	\label{fig:full_weak_comp}
\end{figure}

Early counting approaches either adopt body detection-based techniques~\cite{leibe2005pedestrian} or recur to count mapping~\cite{chan2009bayesian}, which perform unsatisfactorily when encountering highly congested scenes. 
To evade the issue, Lempitsky \etal~\cite{lempitsky2010learning} introduces density maps, thereby casting the counting problem into that of predicting pixel-wise density values. Benefiting from the boom of CNNs, density map prediction has become the mainstream choice in the realm of crowd counting~\cite{zhang2016single,li2018csrnet,liu2019crowd,jiang2020attention,xu2021crowd}; see Figure~\ref{fig:full_weak_comp}(a). 

Albeit the ever-improving counting accuracy, it is notorious that training such pixel-to-pixel probability predictor demands plenty of location-level (strong) annotations. The prevailing datasets with such annotations are generated by manually placing dots at the centroids of people heads across the whole samples (\eg, 1.51 million dots are labelled for the JHU-CROWD++ dataset~\cite{sindagi2020jhu}), which is tedious and time-consuming. Therefore, a variety of recent approaches have been proposed to relieve the labelling burden by employing as few annotated samples as possible.  For example, AL-AC~\cite{zhao2020active} uses limited annotated images via active labelling strategy, whereas other works~\cite{liu2020semi,sindagi2020learning} design semi-supervised schemes to synthesize pseudo ground truths for supervising a larger host of unlabelled images; see Figure~\ref{fig:full_weak_comp}(b). More recently, Xu \etal~\cite{xu2021crowd} make further efforts to reduce the areas of pixel-wise labels by leveraging fixed partial region in each density map; see Figure~\ref{fig:full_weak_comp}(c). But even so, the protocol of using scarce location-level supervision has yet been dispensed with and the requirement of labour-intensive and memory-consuming dot maps persists.

If we rethink the necessity of location-wise spatial constraints, then training crowd counters with count-level annotations, \ie the total number of objects of interest (see Figure~\ref{fig:full_weak_comp}(d)), has many merits. These include: \emph{i}) Training models to predict accurate 2D density maps or dot maps introduces a domain gap between training and inference phases since the models are evaluated solely on total counts, impeding the leaning of object-agnostic repeat patterns;  \emph{ii}) 2D density maps are derived from ground-truth dot maps through a heuristically-defined procedure with inductive bias, which could confuse the models on what to learn; \emph{iii}) Density map requires up to 1,000 times more memory space compared to the global counts, \eg, 362 MB vs. 364 KB for the testing set in ShanghaiTech Part A~\cite{zhang2016single}; and \emph{iv}) More essentially, collecting data with single ground-truth counts can be much easier in many real-world scenarios. For example, when the same scene is captured by multiple cameras, the manual count obtained on one view can serve as ground truths for all captured multi-view images. Furthermore, once an initial count is known for a scene with controlled access, the ground truths for future frames can be obtained easily by adding/subtracting numbers of people entering/leaving. Hence, count-level annotations hold great promises in building large-scale benchmark datasets for multifarious counting problems.

\begin{figure*}[t]
	\begin{center}
		\includegraphics[width=\linewidth]{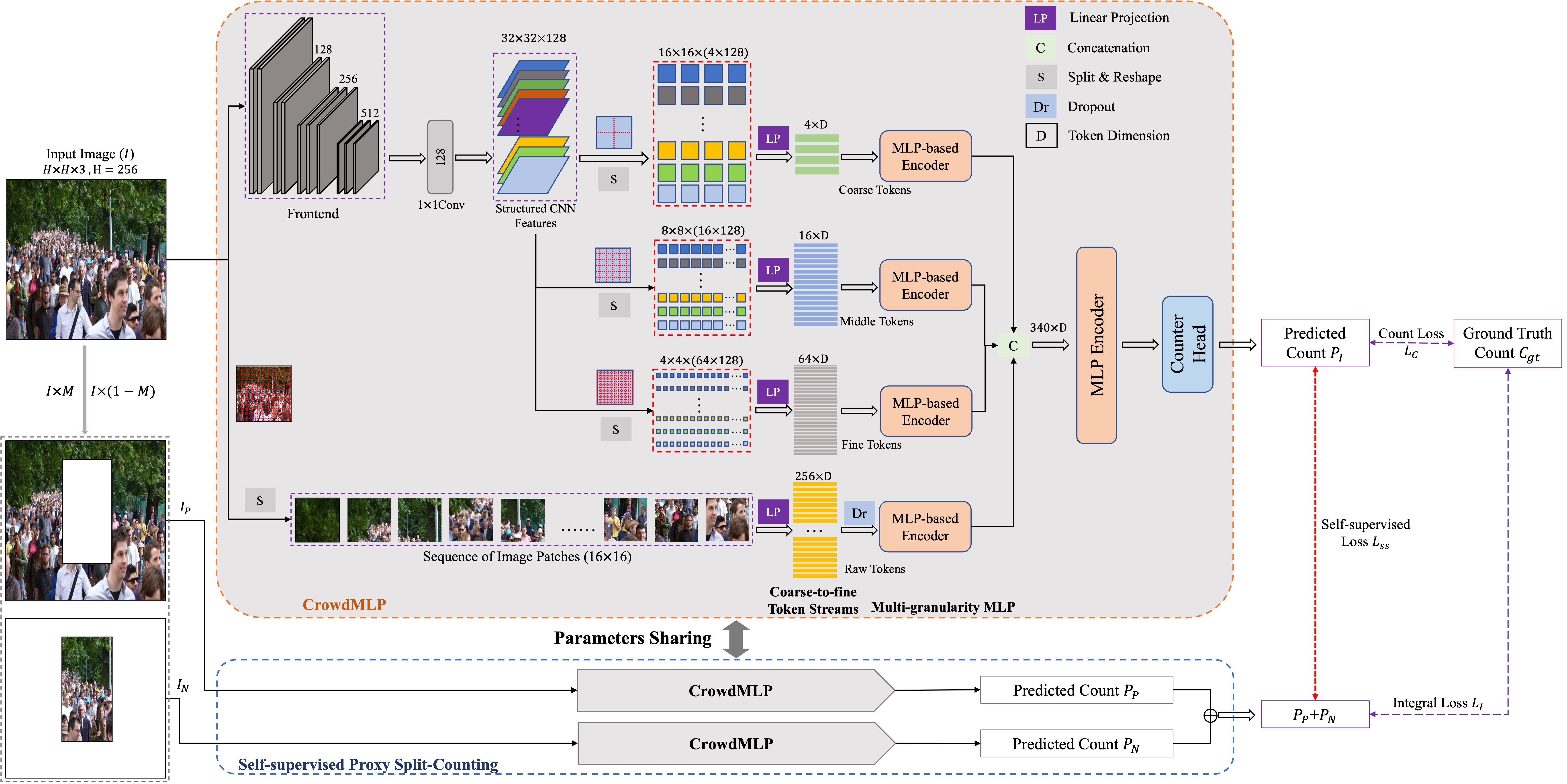}
	\end{center}
	\vspace{-10pt}
	\caption{The overall architecture of CrowdMLP, which is featured by coarse-to-fine token streams, multi-granularity MLP, and self-supervised proxy task.}
	\label{fig:architecture}
\end{figure*}

Recently, a resurge of multi-layer perceptrons (MLP) has lead to cutting-edge purely MLP-based paradigms for image classification~\cite{chen2021cyclemlp,tolstikhin2021mlp}, which benefit from intrinsic advantages of underlying fully-connected layers: more global receptive fields with less inductive bias than \emph{de facto} CNNs component, simpler than self-attention layers in Transformer~\cite{vaswani2017attention}. Inspired by these achievements, this paper presents a novel weakly-supervised {\bf CrowdMLP} architecture, which exploits global-range receptive fields without considering the highly-costly self-attention mechanism and attains better performance than the transformer-based counterpart~\cite{liang2021transcrowd}. As shown in Figure~\ref{fig:architecture}, CrowdMLP is featured by the collaboration between a locally-focused CNN frontend and a devised global-oriented \emph{multi-granularity MLP regressor}. Low-level representations with limited receptive fields are extracted by the pre-trained CNN frontend. These feature maps, along with raw input scenes, are then reshaped and linearly projected to construct a pyramid of token sequences at multiple levels of granularity. The holistic MLP regressor proceed to characterize these tokens and produce embeddings with global-range dependencies, which are fed into a high-level MLP head to directly predict a single and continuous count value.

While training with count-level supervisory signals avoids inductive bias, the lack of location information often causes the performance to degenerate. An auxiliary counting-specific proxy task, namely \emph{Split-Counting},  is designed to stem the degradation by randomly splitting the input into two complementary parts and enforcing the sum counts for the two parts to be close to the count for the whole (see Figure~\ref{fig:architecture}). This strategy implicitly enhance spatial awareness in a self-supervised manner, leading to accuracy gains.

In a nutshell, our contributions are summarized as follows:
\begin{enumerate}
	\item A novel CrowdMLP is proposed to mitigate the burdensome location-level annotations by directly regressing total counts. By proposing multi-granularity MLP regressor with less inductive bias, global-range receptive fields at distinct semantics levels are exploited, while circumventing cumbrous self-attention module in Transformer.

	\item To avoid performance loss caused by the lack of spatial hints in count-level supervision, a self-supervised proxy task is introduced to implicitly impose spatial cues and remedy the accuracy degradation.

	\item Extensive experiments demonstrate that CrowdMLP offers massive performance gains over existing weakly-supervised approaches (\ie CNN- and Transformer-based counterparts) and competitive results as fully-supervised ones. They also shed light on the density- and spatial-awareness properties of CrowdMLP.
\end{enumerate}

\section{Related Work}

\subsection{Counting with Fully-supervised Settings} 

Early approaches tackle the problem of crowd counting by detecting body parts~\cite{leibe2005pedestrian,wu2005detection,viola2005detecting,li2008estimating} or regressing crowd numbers~\cite{chan2009bayesian}. However, when handling highly congested scenes with dense crowd, severe occlusions, and clutter backgrounds, these traditional methods perform poorly. With the impressive capability of CNNs~\cite{simonyan2014very} and the introduction of density map~\cite{lempitsky2010learning}, numerous CNN-based algorithms are proposed to predict density maps.
Among these modern methods, MCNN~\cite{zhang2016single} employs multiple columns with different kernel sizes to extract multi-scale features. Switch-CNN~\cite{babu2017switching} uses an auxiliary classification sub-network to determine which branch should be selected for the regressor at each image patch.
SANet~\cite{cao2018scale} adopts inception module into the feature encoder and presents a local pattern consistency loss. CSRNet~\cite{li2018csrnet} stimulates a line of studies on enlarging the receptive fields via dilated convolutions. CAN~\cite{liu2019context} and SASNet~\cite{song2021choose} investigate the impacts of attention mechanisms on accuracy. Furthermore, BL~\cite{ma2019bayesian}, BM-Count~\cite{liuBipartite2021}, and P2PNet~\cite{song2021rethinking} propose to maximize confidence of head centres with only dot annotation.


\subsection{Counting With Weakly-supervised Settings}
For the sake of reducing the burden of annotations, a suite of models attach importance to weak supervisory signals. 
L2R~\cite{liu2018leveraging} is presented to leverage abundantly-available crowd scenes from the Internet via an auxiliary sorting task with self-supervised settings. Liu \etal~\cite{liu2020semi} design a self-training strategy to train a robust feature extractor from unlabelled samples. Simultaneously, Sindagi \etal~\cite{sindagi2020learning} incorporate a learning mechanism of Gaussian process to generate pseudo labels for unlabelled data. Considering how the model parameters evolve, GWTA-CCNN~\cite{sam2019almost} updates 99.9\% of the parameters in an auto-encoder manner using unlabelled crowd images, whereas remaining weights of the last two layers are fine-tuned with density map supervision. Xu \etal~\cite{xu2021crowd}  use stochastically-selected partial sub-region in each density map to train the model.

Although the above approaches make use of extra unlabelled images or regions, location-level dot or density maps supervisions are still decisive and vital for the final performance.  To get rid of such dependency, Yang \etal~\cite{yang2020weakly} move towards implicitly modelling the hidden relationships in count numbers and predicting the crowd number, which is credited to the side sorting sub-networks and hand-craft soft labels. Nevertheless, the performance achieved is unsatisfactory since CNN-based regressor is incompetent to hunt for global receptive fields. The model is also too sophisticated, making it prone to overfitting. Different from their work, we probe into a model fed by count-level supervisory signals from the perspective of multi-granularity token sequences.

\subsection{Cutting-edge Learning Paradigms}
The past two years witness the focus of preeminent learning paradigms what aim to compete with \emph{de facto} CNNs. Transformer-based ViT~\cite{dosovitskiy2020image} opens a new avenue for treating image patches the similar way as word tokens and attains competitive results on image classification.  Furthermore, the fabric-like models, \ie DETR~\cite{carion2020end}, shows the effectiveness of combining hybrid convolution and transformer. More recently, Tolstikhin \etal~\cite{tolstikhin2021mlp} trigger a line of research to exploit the intrinsic merits of MLPs. Albeit the splendid performance in image classification, the potentials for regression still have not been dived into. {Moreover, multi-head attention mechanism~\cite{vaswani2017attention} has become the standard component in Transformer and intuitively allows for attending to varying parts of the token sequence by passing the input token through the same self-attention unit several times in parallel. Inspired by the merits of multi-head design, we propose the multi-granularity MLP to enrich token embeddings. Considering the drastic scale shifts in the crowd counting, we further refine the plain multi-head method by injecting tokenizations with distinct patch sizes before individual MLP head, which remarkably differs from the multiple attention heads with strictly same inputs and multi-level operations in CNN-based models~\cite{sindagi2019multi,jiang2019learning}.}

\section{Methodology}
As shown in Figure~\ref{fig:architecture}, the proposed count-level framework (CrowdMLP) is comprised of pre-trained CNN frontend, multi-granularity MLP, and self-supervised proxy task.

\subsection{Coarse-to-fine Token Streams}
As aforementioned, global-range receptive fields are pivotal for count-level regression due to the lack of local constraints. To this end, we propose an effective global-oriented MLP regressor which tokenizes the crowd images by breaking it up into a sequence of patches with lower density levels. Apart from the global dependencies, limited training samples and drastic scale variation are prone to impel the model to collapse into trivial outcomes. Hence, a tokenization  structure is designed for marrying the advantages of locally-focused features (pre-trained CNNs) and global-range relationships (MLPs) in a hybrid fashion. In specific, given a square input $I\in \mathcal{R}^{H\times H \times 3}$, where $H\times H$ is image resolution, we first reshape $I$ into a sequence of raw patches of size $16\times 16$, denoted as $R_r$. Meanwhile, following the wide-spread algorithms~\cite{li2018csrnet} for crowd counting, we incorporate first ten layers of a VGG-16 pre-trained on ImageNet as the frontend  to gather crude and low-level features that embrace the implicit spatical hints and the property of translation invariance. The extracted low-level features, with the resolution of $\frac{H}{8}\times \frac{H}{8}$, go through a 1$\times$1 convolution layer to  shorten channel width and suppress model complexity.  The resulted feature maps are denoted as  $C_I$.

To remedy the lack of multi-scale cues in MLPs-based models and increase the cardinality of the model, $C_I$ are split with patch sizes of $16\times 16$, $8\times 8$ and $4\times 4$, respectively, leading to three sets of feature sequences ($F_r$). Then, $F_r$ and $R_r$ are linearly projected to construct four token streams $\{T_1\in \mathcal{R}^{(\frac{H}{8\times16})^2\times D},T_2\in \mathcal{R}^{(\frac{H}{8\times8})^2\times D},T_3\in \mathcal{R}^{(\frac{H}{8\times4})^2\times D},T_4\in \mathcal{R}^{(\frac{H}{16})^2\times D}\}$ at identical token dimension $D$, yielding rich coarse-to-fine tokens for further capturing embeddings with global receptive fields at diverse granularity. The overall process can be defined as follows:
\begin{equation}
\begin{array}{l}
T_1 = LP(S_{16\times 16}(F_r)), T_2 = LP(S_{8\times 8}(F_r))\\
T_3 = LP(S_{4\times 4}(F_r))\\
T_4 = D_r(LP(S_{8\times 8}(R_r))),
\end{array}
\label{equ:pyramid}
\end{equation}
where $S$ denotes split and reshape operations, whereas $LP$ is linear projection. 2D Dropout ($D_r$) with drop rate of 0.2 is employed for raw token sequence to make embeddings robust against noise by stochastically dropping a small set of patches before each epoch.

\subsection{Multi-granularity MLP Regressor}
The MLP-based counter is built upon multiple fundamental transformations, denoted as  $T_{mlp}$, and each of them consists of two plain MLPs (token mixing and channel mixing). Each kind of mixing operator involves two fully-connected layers, non-linear ReLU activation function, and batch normalization $BN$. $T_{mlp}$ can be written as follows:
\begin{equation}
\begin{array}{l}
F_m =ReLU(f_1(w_0, X)),   ~~F_m=Dropout(F_m) \\
F_m=ReLU(f_2(w_1, F_m)), F_m=Dropout(F_m)\\
Y = BN(F_m+X),
\end{array}
\label{equ:mlpmixer}
\end{equation}
where $f_1$ and $f_2$ indicate two plain fully-connected layers, $w_0$ and $w_1$ the corresponding learnable parameters, $X$ and $Y$ the input and output embeddings,  and $F_m$ the intermediate embedding vector.

While token- and channel-mixing communicate information along spatial and channel directions, the interaction across different granularities is not yet considered. This would damage the establishment of complementary embeddings at diverse semantic levels and lead to one-sided long-range dependencies. To attenuate this limitation and mine token embeddings with coarse-to-fine global receptive fields in depth, we present a simple yet effective multi-granularity MLP structure. To the best of our knowledge, this is the first attempt to address the issues of scale/density changes through global-oriented multi-granularity MLP. By imposing this structure on the four input token streams $T_p=\{T_1, T_2, T_3, T_4\}$, our model allows for assembling distinct granularities of token sequences covering varying head scales. Four MLP-based heads $H_1, H_2, H_3, H_4$ with depth of 1 are imposed on $T_p$ according to the Eq.(\ref{equ:mlpmixer}), and generate a pool of embeddings with multi-level global receptive fields, denoted as $E_1, E_2, E_3,E_4$. Sequentially, these higher-level embeddings are concatenated as a holistic embedding vector $\hat{E}$ involving adequate semantics. $\hat{E}$ is then fed into a higher-capable MLP encoder $M_{top}$ with depth of 3 to refine representations and to further mix feature hints at dimensions of cardinality, spatial, and channel. Finally, a light-weigh MLP-based count head $C_{top}$ is plugged at the top of counter to predict the total count $P_I \in R^{1\times 1}$ as a continuous and uncontrolled value. We denote the concatenation as $\odot$ and the overall procedure is depicted as:
\begin{equation}
\small \begin{array}{l}
E_1 = H_1(T_1),~E_2 = H_2(T_2),~E_3 = H_3(T_3),~E_4 = H_4(T_4)\\
\hat{E}=(E_1\odot E_2 \odot E_3 \odot E_4),~ P_I=C_{top}(M_{top}(\hat{E}))
\end{array}
\label{equ:multi-head}
\end{equation}

\subsection{Self-supervised Proxy Split-Counting}
Recently, self-supervised learning is in vogue in computer vision, which has lead to scale on a mass of annotation-free training data and capture robust semantic information. This rule-based learning paradigm provides supervisory signals by pre-designing pivotal surrogate tasks. Towards the goals of decoupling the feature spatial hints and regularizing the learning procedure, we propose a novel proxy task for counting-specific regression, namely \emph{Split-Counting}. It is designed with the consideration of  enforcing the model to capture global-range spatial cues and  regress complementary counts accurately. Specifically, we randomly select a rectangular region of stochastic resolution from input $I$ to form a positive decoupling ($I_P$) and the complement region is referred as a negative decoupling ($I_N$). The generation of pairwise decouplings can be formulated as:
\begin{equation}
I_P=I\times M, I_N=I\times (1-M),
\label{equ:pairwise}
\end{equation}
where  $I = I_N \cup I_P$ and $I_N \cap I_P = \emptyset$. $M$ is randomly-generated mask maps with values of 1 or 0, whereas $\times$ represents pixel-wise multiplication.

The pairwise decouplings ($I_N, I_P$) are taken as the input of our Split-Counting. For simplicity, we denote three lines of regressions as $P_I=M_1(\theta, I)$, $P_P=M_2(\theta, I_P)$, and $P_N=M_3(\theta, I_N)$, where $\theta$ indicates the shared parameters of CrowdMLP and $P$ are the predicted counting results.
By doing so, the intra relationship hidden in a single count value is explicitly decoupled without resorting to extra modules, and the missing spatial constraint is revived. Another key benefit is to reduce regression difficulty by implicitly learning an ensemble ($\hat{M}$) of sub-regressors. 


\subsection{Analysis on Split-Counting}
To theoretically demonstrate the impacts of Split-Counting on lowering empirical error of the entire model, we denote the ensemble regressor as $\hat{M}$ and three implicit sub-regressors as $M_1$, $M_2$ and $M_3$. Hence, we have $\hat{M}=\frac{1}{2}(M_1+(M_2+M_3))$. Following the analysis in DNCL~\cite{zhang2019nonlinear} and bias-variance decomposition theory~\cite{brown2005managing}, given the target crowd count $Y$, the prediction error expectation is decomposed into the combination of bias ``$Bias$", variance ``$Var$" and covariance ``$Cov$" as follows:
\begin{equation}
\begin{array}{l}
E[(\hat{M}-Y)^2] = Bias(\hat{M},Y)^2 + Var(\hat{M})\\
=(E[\hat{M}]-Y)^2 + E[(\hat{M}-E[\hat{M}])^2]\\
=(\frac{1}{2}((E[M_1]-Y)+(E[M_1]+E[M_2]-Y)))^2 \\
+ \frac{1}{4}(E[(M_1-E(M_1))^2]+E[(M_2+M_3-E(M_2+M_3))^2])\\
+Cov(M_1,M_2,M_3)\\
=Bias(\hat{M},Y)^2+Var(\hat{M})+Cov(M_1,M_2,M_3)
\end{array}
\label{equ:mainloss}
\end{equation}
As for the average regression error of each individual regressor, it is straightforward to show:
\begin{equation}
\begin{array}{l}
\frac{1}{2}(E[(M_1-Y)^2]+E[(M_2+M_3-Y)^2])\\
= Bias(\hat{M},Y)^2 + 2Var(\hat{M}) +\frac{1}{2}(Cov(M_1,\hat{M})+Cov(M_2+M_3,\hat{M}))
\end{array}
\label{equ:inter1}
\end{equation}
Then, the variance between sub-individuals and the ensemble model can be transformed as the following equation:
\begin{equation}
\begin{array}{l}
\frac{1}{2}(E[(M_1-\hat{M})^2]+E[(M_2+M_3-\hat{M})^2])\\
= 2Var(\hat{M})-Var(\hat{M})-Cov(M_1,M_2,M_3) \\
+\frac{1}{2}(Cov(M_1,\hat{M})+Cov(M_2+M_3,\hat{M}))
\end{array}
\label{equ:inter2}
\end{equation}
Finally, the regression error of the ensemble model $\hat{M}$ (see equation~(\ref{equ:mainloss})) can be reformulated as the summation of all individual errors and a negative term:
\begin{equation}
\begin{array}{l}
E[(\hat{M}-Y)^2]= \frac{1}{2}(E[(M_1-Y)^2]+E[(M_2+M_3-Y)^2])\\
-\frac{1}{2}(E[(M_1-\hat{M})^2]+E[(M_2+M_3-\hat{M})^2])\\
=\frac{1}{2}(E[(M_1-Y)^2]+E[(M_2+M_3-Y)^2])-E[(M_1-\hat{M})^2]
\end{array}
\label{equ:subloss}
\end{equation}
This indicates that the quadratic error of the ensemble regressor is theoretically guaranteed to be less than or equal to the average of the main baseline and the proxy regressors. Hence, the proposed \emph{Split-Counting} is analytically proven to play a positive role in lowering empirical error of the entire model.

\subsection{Objective Function}
In our framework, we utilize $L_1$ loss between the predicted count $P_I$ and ground-truth count $C_{gt}$ as the pivotal supervisory signal: $L_C = |P_{I}-C_{gt}|$.
In addition, we derive a self-supervised loss ($L_{SS}$) and an integral loss ($L_{I}$) from the proxy task as regularization constraints; see Figure~\ref{fig:architecture}. Given the outputs of the self-supervised Split-Counting, $P_P$ and $P_N$, they are defined as: 
\begin{equation}
L_{SS} = |(P_{P}+P_{N})-P_{I}|, L_{I} = |(P_{P}+P_{N})-C_{gt}|
\label{equ:loss_count}
\end{equation}
The objective function of our CrowdMLP is formulated as $L=L_C+\frac{1}{2}(L_{SS}+L_I)$.

\section{Experiments}

\subsection{Datasets}
Four widely-used benchmark datasets are utilized to verify the proposed method. {\bf ShanghaiTech} dataset~\cite{zhang2016single} consists of two parts: {\bf Part~A} and {\bf Part~B}. This dataset includes 1198 crowd images with total 330,165 annotated people. Classical Part~A contains 482 congested internet images while Part~B consists of 716 sparse images with a fixed size of {768$\times$1024} captured from outside streets.  {\bf UCF\_QNRF}~\cite{idrees2018composition} benchmark is collected from the website and includes 1,553 images with a total number of 1,252,642 people, in which 1201 images are used for training and 334 crowd scenes in testing set. {\bf JHU-CROWD++}~\cite{sindagi2020jhu} is a large-scale crowd dataset including 4,372 images, in which 2,722 images are taken for training, 1,600 scenes for testing, and 500 samples are used for validation. It has a total of 1.51 million annotations on centroids of people and the number of crowd dispersing across crowd scenes ranges from 0 to 25,791. Although NWPU~\cite{gao2020nwpu} is also a popular large-scale dataset, it does not provide the ground truths for testing samples. Hence, the performance on large-scale dataset is evaluated using JUH-CROWD++ here.

\subsection{Implementation Details}
The CrowdMLP is optimized using Adam algorithm with the initial learning rate of 1e-5. The mini-batch size is set to 12 and the learning rate is degraded by a MultiStepLR scheduler. The internal token dimension $D$ is fixed as 256 in our experiments. For all datasets, we first reshape the raw images to keep the max side of 1024 and the other of 768. During the training phase, we randomly crop a batch of patches with the fixed size of 256$\times$256 from the reshaped crowd scenes in an online manner. Stochastic flipping and random lighting with the probability of 0.5 are chosen as data augmentation operations. At the inference time, we design a sliding window strategy with the size of 256$\times$256 to infer a set of sub-counts and their summation is computed as the final count prediction. Two commonly-used metrics, Mean Absolute Error (MAE) and  Mean Square Error (MSE), are employed for performance evaluation: $MAE=\frac{1}{N}\sum_{i=1}^{N}|C_{i}-C_i^{gt}|$ and $MSE=\frac{1}{N}\sum_{i=1}^{N}(C_{i}-C_i^{gt})^2$,
where $N$ represents the number of testing images, $C_i^{gt}$ is the ground-truth total number of people across the $i_{th}$ image and the $C_i$ indicates the corresponding predicted count value.

\subsection{Comparison with State-of-the-art}


As shown in Table~\ref{table:comparision}, on ShanghaiTech Part~A, Part~B, and JHU-CROWD++ datasets, CrowdMLP outperforms the previous state-of-the-art weakly-trained approach (TransCrowd-GAP) significantly, with 19.7\%, 24.7\%, and 13.3\% MSE reductions, respectively. On UCF-QNRF, the improvement is subtle with 3.2\% lower MAE but 1.1\% higher MSE.

As a powerful count-level regressor, CrowdMLP also substantially minimizes the performance gap between count-level and location-level supervised approaches. On ShanghaiTech Part~A, it is slightly behind the best fully-supervised P2PNet (57.828 \emph{vs.} 52.74 on MAE and 84.412 \emph{vs.} 85.06 on MSE), but outperforms other approaches. For example, when compared to the up-to-date BL algorithm, CrowdMLP brings 17.1\% and 4.51\% MAE reduction on ShanghaiTech Part~A and Part~B, respectively. Furthermore, on the large-scale JHU-CROWD++ dataset, which demands more arduous distribution learning, CrowdMLP performs the best among all algorithms that have performances reported to-date. This observation supports our hypothesis that count-level labels provide ample cues for counting, even though they do not explicitly provide spatial information.

\begin{table*}[h]
	\scriptsize
	\begin{center}
		\begin{tabular}{c|c|cc|cc|cc|cc}
			\hline
			\multirow{2}{*}{Methods}&{Label}& \multicolumn{2}{c|}{{\bf Part~A}} 
			&\multicolumn{2}{c|}{{\bf Part~B}}&\multicolumn{2}{c|}{{\bf UCF-QNRF}}&\multicolumn{2}{c}{{\bf JHU++}}\\
			\cline{3-10}
			~& Level & MAE & MSE & MAE & MSE  & MAE & MSE& MAE & MSE\\
			\hline
			ADCrowdNet~\cite{Liu_2019_CVPR}  & Density Map & 63.2 & 98.9 & 7.6 & 13.9  & - & - & - & -\\ 
			PACNN~\cite{Shi_2019_CVPR}  & Density Map & 62.4 & 102.0 & 7.6 & 11.8  & - & - & - & -\\ 
			CAN~\cite{liu2019context}  & Density Map & 62.3 & 100.0 & 7.8 & 12.2 & 107 & 183 & 100.1 & 314.0\\ 
			MBTTBF~\cite{sindagi2019multi} & Density Map & 60.2 & 94.1 & 8.0 & 15.5 & 97.5 & 165.2 & 81.8 & {\bf 299.1}\\ 
			DSSINet~\cite{liu2019crowd} & Density Map & 60.63 &	96.04  & 6.85 &	10.34 & 99.1	&	159.2 & 133.5 & 416.5\\ 
			S-DCNet~\cite{xiong2019open} & Density Map & 58.3 &	95.0  & 6.7	& 10.7 & 104.4	&	176.1& - & - \\ 
			ASNet~\cite{jiang2020attention} & Density Map& 57.78 & 90.13  & -	& - & 91.59	& 159.71 & - & -\\ 
			AMRNet~\cite{liu2020adaptive} &Density Map & 61.59 & 98.36  & 7.02 & 11.00 & 86.6 & 152.2& - & -\\ 
			DM-Count~\cite{wang2020distribution}& Point Map & 59.7 & 95.7  & 7.4 & 11.8 &85.6 & {\bf 148.3} & - & -\\ 
			BL~\cite{ma2019bayesian} & Point Map & 62.8 & 101.8  & 7.7 & 12.7 & 88.7 & 154.8 & {\bf 75.0} & 299.9\\ 
			P2PNet~\cite{song2021rethinking} & Point Map & {\bf 52.74} & {\bf 85.06}   & {\bf 6.25} & {\bf 9.9} & {\bf 85.32} & 154.5& - & -\\ 
			\hline
			MATT~\cite{lei2021towards} & Count+Density &  80.1 & 129.4   &    11.7 & 17.5 & - & -& - & -\\
			Sorting~\cite{yang2020weakly} & Count & 104.6 & 145.2   &  12.3 & 21.2 & - & -& - & -\\
			TransCrowd-T~\cite{liang2021transcrowd} & Count &  69.0 & 116.5   &   10.6 & 19.7 & 98.9 & 176.1& 76.4 & 319.8\\
			TransCrowd-G~\cite{liang2021transcrowd} & Count &  66.1 & 105.1   &   9.3 & 16.1 & 97.2 & {\bf 168.5}& 74.9 & 295.6\\
			{\bf CrowdMLP} (ours) & Count & {\bf 57.828} & {\bf 84.412} & {\bf 7.667} & {\bf 12.127}  & {\bf 94.130} & {170.324} & {\bf 67.568} & {\bf 256.154}\\
			\hline
		\end{tabular}
	\end{center}
	\caption{
		Quantitative comparisons with state-of-the-art approaches under different annotation levels on four datasets. Best results under each annotation level are shown in boldface. Our approach significantly outperforms existing count-level approaches.  It also performs better than most location-level approaches on the ShanghaiTech Part~A dataset, while attaining the best reported accuracy on JHU-CROWD++.
	}
	\label{table:comparision}
\end{table*}


\subsection{Ablation Study}
Ablation studies are conducted on ShanghaiTech Part A dataset to thoroughly investigate the effects of each component. For simplicity, the presented CrowdMLP is denoted as the Baseline.

{\bf The Effects of Coarse-to-fine Token Streams.} We first ablate the effects of structuring token sequences at multiple levels of granularity by removing each sub-encoder. Table~\ref{table:ab_head} shows that removing individual {encoders} from the pyramid incurs performance loss, which confirms the importance of the devised multi-granularity MLP.

\begin{table}[h]
	\begin{center}
		\setlength{\tabcolsep}{3mm}
		\small
		\begin{tabular}{c|c|cc}
			\hline
			\multirow{2}{*}{Methods}&	\multirow{2}{*}{Parameters}& \multicolumn{2}{c}{{Part~A}}\\
			\cline{3-4}
			~& & MAE & MSE\\
			\hline
			 w.o. Raw Token & 23.70M & 60.411 & 88.613\\
			 w.o. 16$\times$16 Token & 17.65M & 61.896 & 93.894\\
			 w.o. 8$\times$8 Token & 23.74M & 61.697 & 87.717\\
			 w.o. 4$\times$4 Token & 24.56M & 59.294 & 87.483\\
			 Baseline & 26.63M & 57.828 & 84.412\\
			\hline
		\end{tabular}
	\end{center}
	\caption{Ablation study on multi-granularity structure. The first row removes the token sequence for raw image patches from the  coarse-to-fine token streams, whereas each of the following 3 rows removes those for feature patches of corresponding sizes. Performance degradation is observed when any token sequence is removed.}
	\label{table:ab_head}
\end{table}

{\bf Using Different Learning Paradigms.} 
Next, we compare the regression capability of MLPs-based structure against CNNs and Transformer paradigms. A CSRNet-like CNNs-based model (Crowd-CNN) is constructed for directly counting under the count-level supervision. For fair comparison, the width and depth of Crowd-CNN is carefully adjusted to match the number of parameters in the baseline. A second variant (Crowd-Transformer) is obtained by replacing all MLPs transformations in the baseline with transformer in ViT that have the same depth and dimension.

Table~\ref{table:ab_mlp} shows that Crowd-CNN delivers a seriously degraded performance, which may be attributed to the fact that CNNs unit is incapable of modelling long-range receptive fields while focusing on locality~\cite{yang2020weakly}. The lower errors achieved by Crowd-Transformer and CrowdMLP verify the importance of global-range receptive fields for count-level regressors. Albeit the acceptable accuracy, Crowd-Transformer suffers from increased model scale (36.43M \emph{vs.} 26.63M) and high computational complexity. Our observation is consistent with the recent $S^2$-MLPv2~\cite{yu2021s} for image classification.

\begin{table}[h]
	\begin{center}
		\setlength{\tabcolsep}{2.5mm}
		\small
		\begin{tabular}{c|c|cc}
			\hline
			\multirow{2}{*}{Methods}&	\multirow{2}{*}{Parameters}& \multicolumn{2}{c}{{Part~A}}\\
			\cline{3-4}
			~& & MAE & MSE\\
			\hline
			Crowd-CNN & 26.63M & 159.925 & 251.649\\
			Crowd-Transformer & 36.43M & 61.121 & 92.922\\
			Baseline & 26.63M & 57.828 & 84.412\\
			\hline
		\end{tabular}
	\end{center}
	\caption{The effects of different regression paradigms involving CNNs, Transformers, and MLPs.
	}
	\label{table:ab_mlp}
\end{table}

\begin{figure}[ht]
	\begin{center}
		\includegraphics[width=0.8\linewidth]{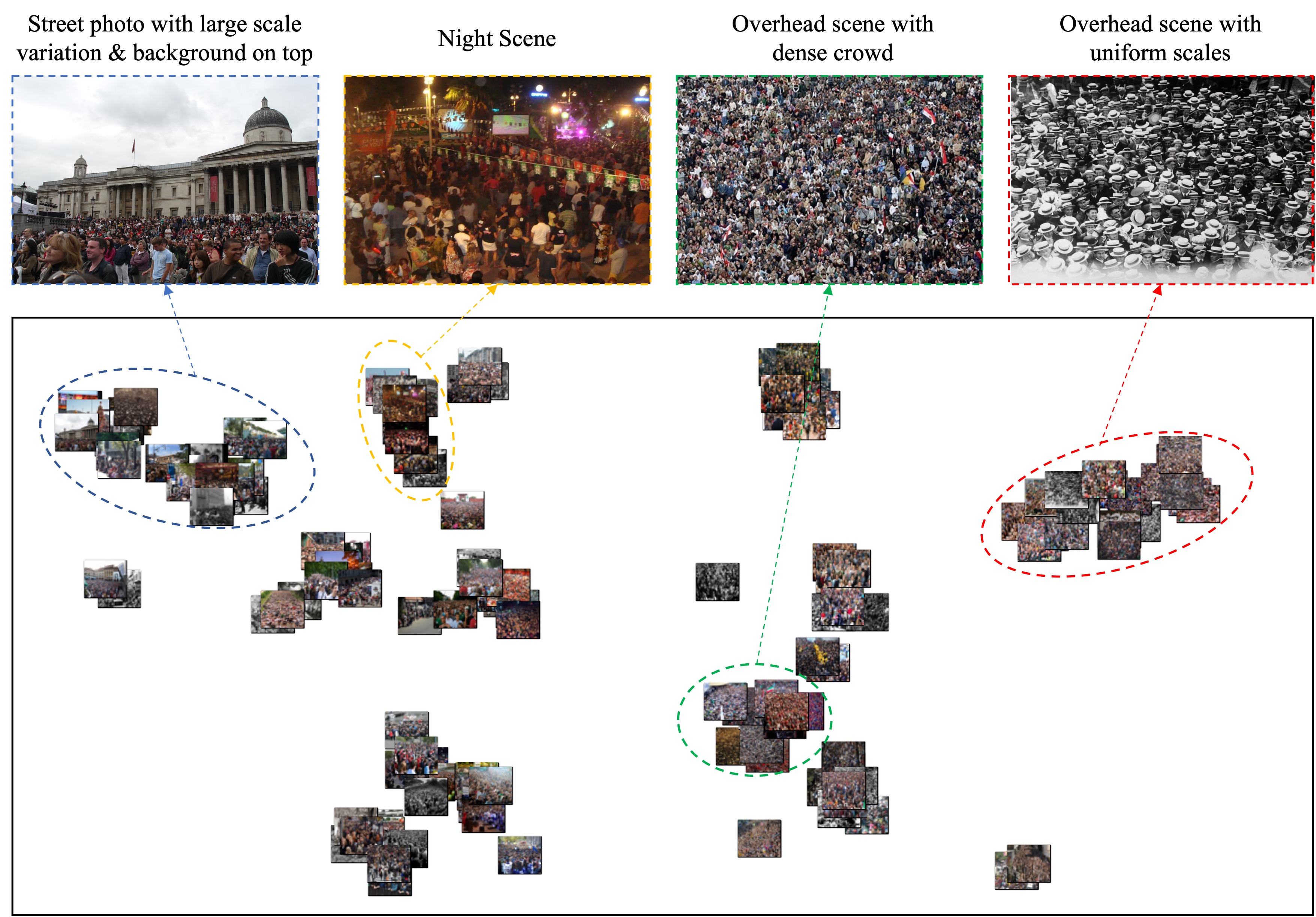}
	\end{center}
	\vspace{-10pt}
	\caption{Cluster visualization of testing scenes in ShanghaiTech Part~A by imposing T-SNE on learned intermediate features. Scenes with varying attributes fall into different clusters. From left to right, the zoomed-in views show examples from selected clusters.}
	\label{fig:feature_cluster}
\end{figure}

\begin{figure}[h]
	\begin{center}
		\includegraphics[width=0.75\linewidth]{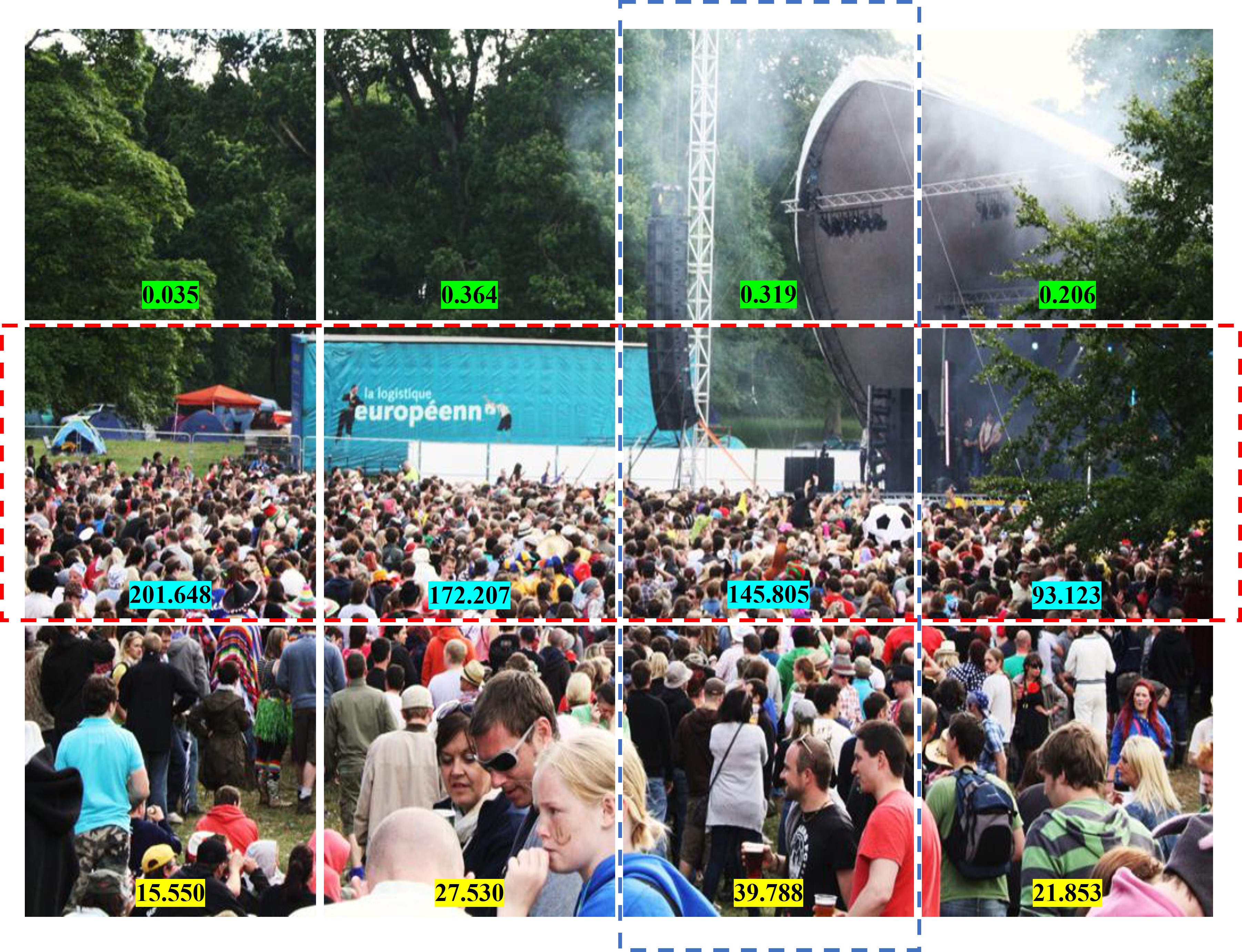}
	\end{center}
	\caption{Visualization on sliding window counting results on a crowd scene example with drastic scale, density and semantics shifts across the whole image. The total predicted count is 718.4, whereas the ground truth is 717.}
	\label{fig:spatial_example}
\end{figure}

\begin{figure}[ht]
	\begin{center}
		\includegraphics[width=0.8\linewidth]{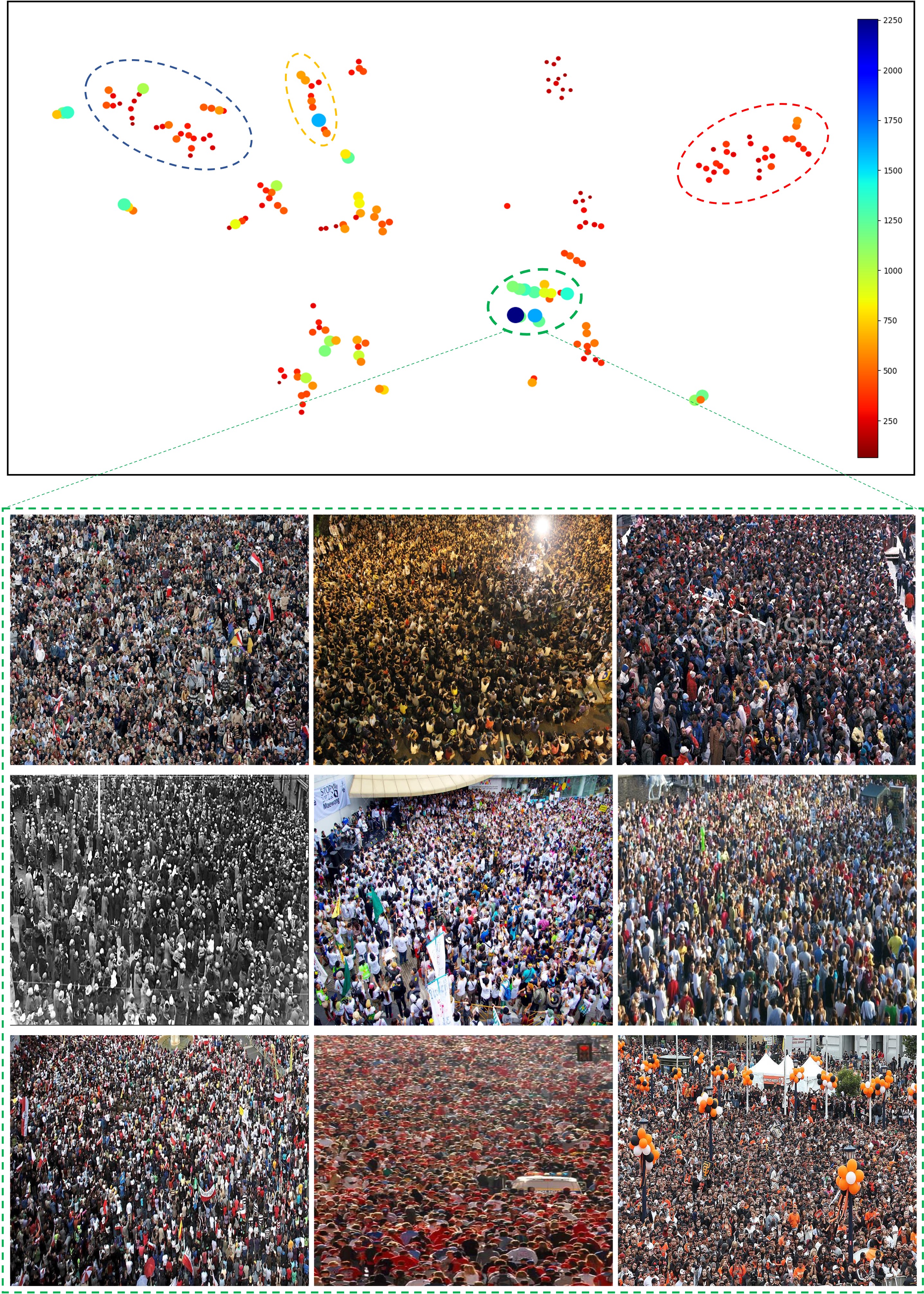}
	\end{center}
	\vspace{-10pt}
	\caption{Embeddings cluster visualization of testing images in ShanghaiTech Part A using T-SNE technique. Each image is shown as a colour dot, where the colour represents the crowd number/densities. While the images are clustered primarily by scene features (see Figure 3), high-density images tend to group together (green cluster). The zoomed-in view depicts corresponding real scenes in green cluster.}
	\label{fig:featurecluster}
\end{figure}

{\bf The Effect of Proxy Split-Counting.} To evaluate the impacts of the proxy task, we also train a variant of CrowdMLP without the self-supervised \emph{Split-Counting} on ShanghaiTech Part~A. The comparison shows that Split-Counting helps to bring 0.55\% improvement on MAE and 0.32\% improvement on MSE. This confirms our theoretical analysis that the proxy task is beneficial for implicitly decoupling spatial constraints and lowering the empirical error. 


\subsection{Visualization and Analysis}
To discern learned features in depth under the count-level supervisions and investigate whether or not CrowdMLP, trained without any location signals, is aware of scale variation, density shift, and spatial semantics, we visualize the embeddings using T-SNE technique and present a typical inference example with drastic scale/density/semantics changes across the whole image. The clustering results in Figures~\ref{fig:feature_cluster} and \ref{fig:featurecluster} show that CrowdMLP learns discriminative features with underlying repetitive patterns for diverse scenes. Photos with similar properties appear to be grouped into same clusters. For example, on the leftmost (blue) cluster, we have street photos with large scale variation where crowd only show up at the bottom of the images, whereas on the rightmost (red) cluster, we have overhead views with uniformly distributed crowd.

Simultaneously, Figure~\ref{fig:spatial_example} exhibits the behaviour of sliding-window strategy with the size of 256$\times$256 at the testing phase. It can be observed that, although lacking location-level prior, our weakly-supervised CrowdMLP can distinguish spatial semantics (the background and foreground people). For the common issues of scale variations and density shifts, our model seems to work pretty well and predict people number precisely under the variation trend of scale and density; see red and blue dotted regions in the Figure~\ref{fig:spatial_example}.

\subsection{Conclusion}

In this paper, a novel weakly-supervised CrowdMLP is presented to mitigate the burden of the labour-intensive location-level annotations and learn to directly regress crowd counts using only count-level labels. In specific, the multi-granularity MLP regressor are creatively designed to capture global receptive fields at varying levels of granularity. Besides, a promising proxy task, namely split-counting, is introduced for regularizing the learning procedure of the crowd regressor, implicitly decoupling spatial constraints, and avoiding performance degradation. The split-counting also brings potentials for exploiting massive unlabelled or cross-domain data.  Extensive experiments have demonstrated the effectiveness of the presented method. For future work, we plan to explore the potential of CrowdMLP for cross-domain counting and/or domain adaptation.

%
%
\bibliographystyle{splncs04}
\bibliography{egbib}

\begin{thebibliography}{10}
\providecommand{\url}[1]{\texttt{#1}}
\providecommand{\urlprefix}{URL }
\providecommand{\doi}[1]{https://doi.org/#1}

\bibitem{babu2017switching}
Babu~Sam, D., Surya, S., Venkatesh~Babu, R.: Switching convolutional neural
  network for crowd counting. In: Proceedings of the IEEE conference on
  computer vision and pattern recognition. pp. 5744--5752 (2017)

\bibitem{brown2005managing}
Brown, G., Wyatt, J.L., Tino, P., Bengio, Y.: Managing diversity in regression
  ensembles. Journal of machine learning research  \textbf{6}(9) (2005)

\bibitem{cao2018scale}
Cao, X., Wang, Z., Zhao, Y., Su, F.: Scale aggregation network for accurate and
  efficient crowd counting. In: Proceedings of the European conference on
  computer vision (ECCV). pp. 734--750 (2018)

\bibitem{carion2020end}
Carion, N., Massa, F., Synnaeve, G., Usunier, N., Kirillov, A., Zagoruyko, S.:
  End-to-end object detection with transformers. In: European conference on
  computer vision. pp. 213--229. Springer (2020)

\bibitem{chan2009bayesian}
Chan, A.B., Vasconcelos, N.: Bayesian poisson regression for crowd counting.
  In: 2009 IEEE 12th international conference on computer vision. pp. 545--551.
  IEEE (2009)

\bibitem{chen2021cyclemlp}
Chen, S., Xie, E., et~al.: Cyclemlp: A mlp-like architecture for dense
  prediction. arXiv preprint arXiv:2107.10224  (2021)

\bibitem{dosovitskiy2020image}
Dosovitskiy, A., Beyer, L., Kolesnikov, A., Weissenborn, D., Zhai, X.,
  Unterthiner, T., Dehghani, M., Minderer, M., Heigold, G., Gelly, S., et~al.:
  An image is worth 16x16 words: Transformers for image recognition at scale.
  arXiv preprint arXiv:2010.11929  (2020)

\bibitem{idrees2018composition}
Idrees, H., Tayyab, M., Athrey, K., Zhang, D., Al-Maadeed, S., Rajpoot, N.,
  Shah, M.: Composition loss for counting, density map estimation and
  localization in dense crowds. In: Proceedings of the european conference on
  computer vision (ECCV). pp. 532--546 (2018)

\bibitem{jiang2019learning}
Jiang, X., Zhang, L., Lv, P., Guo, Y., Zhu, R., Li, Y., Pang, Y., Li, X., Zhou,
  B., Xu, M.: Learning multi-level density maps for crowd counting. IEEE
  transactions on neural networks and learning systems  \textbf{31}(8),
  2705--2715 (2019)

\bibitem{jiang2020attention}
Jiang, X., Zhang, L., Xu, M., Zhang, T., Lv, P., Zhou, B., Yang, X., Pang, Y.:
  Attention scaling for crowd counting. In: Proceedings of the IEEE/CVF
  Conference on Computer Vision and Pattern Recognition. pp. 4706--4715 (2020)

\bibitem{lei2021towards}
Lei, Y., Liu, Y., et~al.: Towards using count-level weak supervision for crowd
  counting. PR  (2021)

\bibitem{leibe2005pedestrian}
Leibe, B., Seemann, E., Schiele, B.: Pedestrian detection in crowded scenes.
  In: 2005 IEEE Computer Society Conference on Computer Vision and Pattern
  Recognition (CVPR'05). vol.~1, pp. 878--885. IEEE (2005)

\bibitem{lempitsky2010learning}
Lempitsky, V., Zisserman, A.: Learning to count objects in images. Advances in
  neural information processing systems  \textbf{23} (2010)

\bibitem{li2008estimating}
Li, M., Zhang, Z., Huang, K., Tan, T.: Estimating the number of people in
  crowded scenes by mid based foreground segmentation and head-shoulder
  detection. In: 2008 19th international conference on pattern recognition.
  pp.~1--4. IEEE (2008)

\bibitem{li2018csrnet}
Li, Y., Zhang, X., Chen, D.: Csrnet: Dilated convolutional neural networks for
  understanding the highly congested scenes. In: Proceedings of the IEEE
  conference on computer vision and pattern recognition. pp. 1091--1100 (2018)

\bibitem{liang2021transcrowd}
Liang, D., Chen, X., et~al.: Transcrowd: Weakly-supervised crowd counting with
  transformer. arXiv preprint arXiv:2104.09116  (2021)

\bibitem{liuBipartite2021}
Liu, H., Zhao, Q., Ma, Y., Dai, F.: Bipartite matching for crowd counting with
  point supervision. In: International Joint Conference on Artificial
  Intelligence (2021)

\bibitem{liu2019crowd}
Liu, L., Qiu, Z., Li, G., Liu, S., Ouyang, W., Lin, L.: Crowd counting with
  deep structured scale integration network. In: Proceedings of the IEEE/CVF
  international conference on computer vision. pp. 1774--1783 (2019)

\bibitem{Liu_2019_CVPR}
Liu, N., Long, Y., Zou, C., Niu, Q., Pan, L., Wu, H.: Adcrowdnet: An
  attention-injective deformable convolutional network for crowd understanding.
  In: Proceedings of the IEEE/CVF Conference on Computer Vision and Pattern
  Recognition. pp. 3225--3234 (2019)

\bibitem{liu2019context}
Liu, W., Salzmann, M., Fua, P.: Context-aware crowd counting. In: Proceedings
  of the IEEE/CVF Conference on Computer Vision and Pattern Recognition. pp.
  5099--5108 (2019)

\bibitem{liu2018leveraging}
Liu, X., Van De~Weijer, J., Bagdanov, A.D.: Leveraging unlabeled data for crowd
  counting by learning to rank. In: Proceedings of the IEEE conference on
  computer vision and pattern recognition. pp. 7661--7669 (2018)

\bibitem{liu2020adaptive}
Liu, X., Yang, J., Ding, W., Wang, T., Wang, Z., Xiong, J.: Adaptive mixture
  regression network with local counting map for crowd counting. In: European
  Conference on Computer Vision. pp. 241--257. Springer (2020)

\bibitem{liu2020semi}
Liu, Y., Liu, L., Wang, P., Zhang, P., Lei, Y.: Semi-supervised crowd counting
  via self-training on surrogate tasks. In: European Conference on Computer
  Vision. pp. 242--259. Springer (2020)

\bibitem{lu2017tasselnet}
Lu, H., Cao, Z., Xiao, Y., Zhuang, B., Shen, C.: Tasselnet: counting maize
  tassels in the wild via local counts regression network. Plant methods
  \textbf{13}(1),  1--17 (2017)

\bibitem{ma2021spatiotemporal}
Ma, Y.J., Shuai, H.H., Cheng, W.H.: Spatiotemporal dilated convolution with
  uncertain matching for video-based crowd estimation. IEEE Transactions on
  Multimedia  (2021)

\bibitem{ma2021towards}
Ma, Z., Hong, X., Wei, X., Qiu, Y., Gong, Y.: Towards a universal model for
  cross-dataset crowd counting. In: Proceedings of the IEEE/CVF International
  Conference on Computer Vision. pp. 3205--3214 (2021)

\bibitem{ma2019bayesian}
Ma, Z., Wei, X., Hong, X., Gong, Y.: Bayesian loss for crowd count estimation
  with point supervision. In: Proceedings of the IEEE/CVF International
  Conference on Computer Vision. pp. 6142--6151 (2019)

\bibitem{punn2020monitoring}
Punn, N.S., Sonbhadra, S.K., Agarwal, S., Rai, G.: Monitoring covid-19 social
  distancing with person detection and tracking via fine-tuned yolo v3 and
  deepsort techniques. arXiv preprint arXiv:2005.01385  (2020)

\bibitem{sam2019almost}
Sam, D.B., Sajjan, N.N., Maurya, H., Babu, R.V.: Almost unsupervised learning
  for dense crowd counting. In: Proceedings of the AAAI Conference on
  Artificial Intelligence. vol.~33, pp. 8868--8875 (2019)

\bibitem{Shi_2019_CVPR}
Shi, M., Yang, Z., Xu, C., Chen, Q.: Revisiting perspective information for
  efficient crowd counting. In: Proceedings of the IEEE/CVF Conference on
  Computer Vision and Pattern Recognition. pp. 7279--7288 (2019)

\bibitem{simonyan2014very}
Simonyan, K., Zisserman, A.: Very deep convolutional networks for large-scale
  image recognition. arXiv preprint arXiv:1409.1556  (2014)

\bibitem{sindagi2020jhu}
Sindagi, V., Yasarla, R., Patel, V.M.: Jhu-crowd++: Large-scale crowd counting
  dataset and a benchmark method. IEEE Transactions on Pattern Analysis and
  Machine Intelligence  (2020)

\bibitem{sindagi2019multi}
Sindagi, V.A., Patel, V.M.: Multi-level bottom-top and top-bottom feature
  fusion for crowd counting. In: Proceedings of the IEEE/CVF international
  conference on computer vision. pp. 1002--1012 (2019)

\bibitem{sindagi2020learning}
Sindagi, V.A., Yasarla, R., Babu, D.S., Babu, R.V., Patel, V.M.: Learning to
  count in the crowd from limited labeled data. In: European Conference on
  Computer Vision. pp. 212--229. Springer (2020)

\bibitem{song2021rethinking}
Song, Q., Wang, C., Jiang, Z., Wang, Y., Tai, Y., Wang, C., Li, J., Huang, F.,
  Wu, Y.: Rethinking counting and localization in crowds: A purely point-based
  framework. In: Proceedings of the IEEE/CVF International Conference on
  Computer Vision. pp. 3365--3374 (2021)

\bibitem{song2021choose}
Song, Q., Wang, C., Wang, Y., Tai, Y., Wang, C., Li, J., Wu, J., Ma, J.: To
  choose or to fuse? scale selection for crowd counting. In: Proceedings of the
  AAAI Conference on Artificial Intelligence. vol.~35, pp. 2576--2583 (2021)

\bibitem{tolstikhin2021mlp}
Tolstikhin, I.O., Houlsby, N., Kolesnikov, A., Beyer, L., Zhai, X.,
  Unterthiner, T., Yung, J., Steiner, A., Keysers, D., Uszkoreit, J., et~al.:
  Mlp-mixer: An all-mlp architecture for vision. Advances in Neural Information
  Processing Systems  \textbf{34} (2021)

\bibitem{vaswani2017attention}
Vaswani, A., Shazeer, N., Parmar, N., Uszkoreit, J., Jones, L., Gomez, A.N.,
  Kaiser, {\L}., Polosukhin, I.: Attention is all you need. Advances in neural
  information processing systems  \textbf{30} (2017)

\bibitem{viola2005detecting}
Viola, P., Jones, M.J., Snow, D.: Detecting pedestrians using patterns of
  motion and appearance. International Journal of Computer Vision
  \textbf{63}(2),  153--161 (2005)

\bibitem{wang2020distribution}
Wang, B., Liu, H., Samaras, D., Nguyen, M.H.: Distribution matching for crowd
  counting. Advances in Neural Information Processing Systems  \textbf{33},
  1595--1607 (2020)

\bibitem{gao2020nwpu}
Wang, Q., Gao, J., Lin, W., Li, X.: Nwpu-crowd: A large-scale benchmark for
  crowd counting and localization. IEEE Transactions on Pattern Analysis and
  Machine Intelligence  (2020). \doi{10.1109/TPAMI.2020.3013269}

\bibitem{wu2005detection}
Wu, B., Nevatia, R.: Detection of multiple, partially occluded humans in a
  single image by bayesian combination of edgelet part detectors. In: Tenth
  IEEE International Conference on Computer Vision (ICCV'05) Volume 1. vol.~1,
  pp. 90--97. IEEE (2005)

\bibitem{xiong2019open}
Xiong, H., Lu, H., Liu, C., Liu, L., Cao, Z., Shen, C.: From open set to closed
  set: Counting objects by spatial divide-and-conquer. In: Proceedings of the
  IEEE/CVF International Conference on Computer Vision. pp. 8362--8371 (2019)

\bibitem{xu2021crowd}
Xu, Y., Zhong, Z., Lian, D., Li, J., Li, Z., Xu, X., Gao, S.: Crowd counting
  with partial annotations in an image. In: Proceedings of the IEEE/CVF
  International Conference on Computer Vision. pp. 15570--15579 (2021)

\bibitem{yang2020weakly}
Yang, Y., Li, G., Wu, Z., Su, L., Huang, Q., Sebe, N.: Weakly-supervised crowd
  counting learns from sorting rather than locations. In: European Conference
  on Computer Vision. pp. 1--17. Springer (2020)

\bibitem{yu2021s}
Yu, T., Li, X., et~al.: S2mlpv2: Improved spatial-shift mlp architecture for
  vision. arXiv preprint arXiv:2108.01072  (2021)

\bibitem{zhang2019nonlinear}
Zhang, L., Shi, Z., Cheng, M.M., Liu, Y., Bian, J.W., Zhou, J.T., Zheng, G.,
  Zeng, Z.: Nonlinear regression via deep negative correlation learning. IEEE
  transactions on pattern analysis and machine intelligence  \textbf{43}(3),
  982--998 (2019)

\bibitem{Zhang_2017_ICCV}
Zhang, S., Wu, G., Costeira, J.P., Moura, J.M.: Fcn-rlstm: Deep spatio-temporal
  neural networks for vehicle counting in city cameras. In: Proceedings of the
  IEEE international conference on computer vision. pp. 3667--3676 (2017)

\bibitem{zhang2016single}
Zhang, Y., Zhou, D., Chen, S., Gao, S., Ma, Y.: Single-image crowd counting via
  multi-column convolutional neural network. In: Proceedings of the IEEE
  conference on computer vision and pattern recognition. pp. 589--597 (2016)

\bibitem{zhao2020active}
Zhao, Z., Shi, M., Zhao, X., Li, L.: Active crowd counting with limited
  supervision. In: European Conference on Computer Vision. pp. 565--581.
  Springer (2020)

\end{thebibliography}
\end{document}